\documentclass{llncs}

\usepackage[a4paper,includeheadfoot,top=1.65in,bottom=1.65in,left=1.73in,hcentering]{geometry}
\usepackage[pdftex]{graphicx}
\usepackage[sectionbib]{natbib}
\usepackage{booktabs}
\usepackage{caption} 
\usepackage{multirow}
\usepackage{amsmath}

\usepackage{tablefootnote}

\usepackage[utf8]{inputenc}
\usepackage{t1enc}
\usepackage{tikz-dependency}
\usepackage{qtree}
\usepackage{float}
\restylefloat{figure}
\usepackage{tikz-qtree}
\usepackage{booktabs}
\newif{\ifhidecomments}
\hidecommentsfalse

\ifhidecomments
    \newcommand{\botond}[1]{}
    \newcommand{\attila}[1]{}
    \newcommand{\judit}[1]{}
    \newcommand{\dorina}[1]{}
\else
    \newcommand{\botond}[1]{\textcolor{red}{[#1 ({\bf Botond})]}}
    \newcommand{\attila}[1]{\textcolor{blue}{[#1 ({\bf Attila})]}} 
    \newcommand{\judit}[1]{\textcolor{orange}{[#1 ({\bf Judit})]}} 
    \newcommand{\dorina}[1]{\textcolor{green}{[#1 ({\bf Dorina})]}}
\fi

\begin{document}

\title{HunSum-1: an Abstractive Summarization Dataset for Hungarian}
\author{Botond Barta\inst{1,2}, Dorina Lakatos\inst{1,2}, Attila Nagy\inst{1}, Mil\'an Konor Nyist\inst{1}, Judit \'Acs\inst{2}\\
\institute{
$^1$Department of Automation and Applied Informatics\\
Budapest University of Technology and Economics\\
$^2$Institute for Computer Science and Control\\
Eötvös Loránd Research Network}
\email{\{botondbarta, dorinalakatos, acsjudit\}@sztaki.hu}, \email{attila.nagy234@gmail.com}, \email{nyist.milan78@gmail.com}
}

\maketitle

\begin{abstract}
We introduce \textit{HunSum-1}: a dataset for Hungarian abstractive summarization, consisting of 1.14M news articles. The dataset is built by collecting, cleaning and deduplicating data from 9 major Hungarian news sites through CommonCrawl. Using this dataset, we build abstractive summarizer models based on huBERT and mT5. We demonstrate the value of the created dataset by performing a quantitative and qualitative analysis on the models' results. The \textit{HunSum-1} dataset, all models used in our experiments and our code\footnote{https://github.com/dorinapetra/summarization} are available open source.
\end{abstract}

\section{Introduction}
Automatic summarization of documents \citep{nenkova-etal-2011-automatic} is a widely studied task in natural language processing. The goal of summarization is to reduce the original document to a short, concise text, such that it captures every key information from the input. Generally, there are two approaches to summarization: extractive and abstractive. Extractive methods directly use the input document to produce the summary by extracting relevant tokens or sentences explicitly. 
In contrast, abstractive summarization generates new text conditioned on the input.
Extractive summarization methods are usually less complex from a computational perspective compared to their abstractive counterparts, however, they have limitations such as handling long sentences or ensuring coherence among the extracted sentences. Producing more sophisticated summaries requires abstractive summarization, where the model may generate words, which are not present in the original document.

In this paper, we introduce \textit{HunSum-1}, an abstractive summarization dataset for Hungarian consisting of 1.14M news articles with leads. In Section~\ref{sec:dataset} we discuss in detail how we extracted the data from CommonCrawl and performed a number of preprocessing steps to arrive at the current version of \textit{HunSum-1}. We also train three baseline models on the dataset and evaluate the results both quantitatively and qualitatively. To the best of our knowledge, this is the first publicly available Hungarian abstractive summarization dataset of this size. We hope this will serve as a good foundation for future abstractive summarization research in Hungarian.

\section{Related work}

% Draft
% TODO: Summarize key papers in 1-2 sentences, but mostly focus on datasets and Hungarian aspects since this is our focus.

% Extractive summarization key papers
\cite{summarunner} introduce an RNN based extractive summarization model called \textit{SummaRuNNer}, which outperformed several models, such as a feature-rich logistic classifier \citep{DBLP:journals/corr/ChengL16a}, an Integer Linear Programming based approach \citep{woodsend-lapata-2010-automatic} and various graph based models \citep{parveen-etal-2015-topical, wan2010towards}. Traditionally there are two main steps in an extractive summarization model: sentence scoring and sentence selection.
\citet{Neusum} integrate the two steps and directly predict the importance of the sentence based on the previous ones.
\citet{bert_extractive} introduces \textsc{Bertsum}, an end-to-end BERT-based \citep{bert} model with inter-sentence Transformer layers.
\textsc{Bertsum} achieves better ROUGE-L scores than \textsc{Neusum}, the previous state-of-the-art model on the \textit{CNN/Daily Mail} \citep{NIPS2015_afdec700} dataset. 

% Abstractive summarization key papers
For abstractive summarization, \cite{DBLP:journals/corr/PaulusXS17} introduce an encoder-decoder based neural intra-attention model and a new training procedure that combines supervised and reinforcement learning. \cite{DBLP:journals/corr/abs-1808-10792} propose the \textit{bottom-up} attention, a new technique for content selection: it uses a data-efficient content selector to over-determine the important phrases from the input. The BERT-based \textsc{PreSumm} model \citep{presumm} uses a new fine-tuning schedule which adopts different optimizers for the encoder and the decoder. The current state-of-the-art model, PEGASUS was introduced by \cite{pmlr-v119-zhang20ae}. During the training of PEGASUS, the important sentences are masked from the input document, and the models have to generate these sentences together as one output.
This training objective eliminates the need for a supervised summarization dataset.

% Models (e.g. mT5 for summarization)

\cite{t5} introduce T5 (Text-To-Text Transfer Transformer) a Transformer based encoder-decoder which was pretrained on several downstream tasks including summarization, but only for English. mT5 \citep{mt5} is the multilingual version of the T5 model.
\cite{xlsum} fine-tune the mT5 model for multilingual summarization and achieve competitive results compared to monolingual models.
% Judit: I'm removing ParsBERT. If we mention language specific models, we should mention more than one
%\cite{persian_mt5} introduce two abstractive summarization models for Persian: an encoder-decoder version of the ParsBERT model \citep{ParsBERT} and a fine-tuned mT5 model.

% Other datasets (XL-sum has Hungarian for example, need to talk about that)
\cite{NIPS2015_afdec700} create a dataset from pairing the summaries with CNN and Daily Mail articles. The \textit{CNN/Dailymail} summarization corpus contains over 1 million samples.
\cite{xlsum} introduce a similar dataset called \textit{XL-Sum}, which has 1 million annotated data samples in 44 languages. They extracted the lead as the summary and the article as the input document from BBC news. 

% Mention & cite every paper that was published on Hungarian summarization (also might want to check related work section of these papers)
\cite{yang} build a summarization corpus from Index and HVG articles and train several BERT models for abstractive summarization: multilingual cased BERT, huBERT \citep{Nemeskey:2020} and the HILBERT large models \citep{feldmannhilbert}.
The corpus is unfortunately not publicly available.
\cite{makraitowards} publish an encoder-decoder based model initialized with huBERT. They use the ELTE.DH corpus  with 300k samples for training. \cite{presumm_yang} used the PreSumm method for training multilingual and Hungarian BERT models. \cite{barterezzunk} finetunes BART models \citep{lewis-etal-2020-bart} for abstractive summarization.
%http://publikacio.uni-eszterhazy.hu/7012/1/AMI_53_from299to316.pdf
%https://hlt.bme.hu/media/pdf/fosztogatnak2osztogatnak_lZ3ji2C.pdf

%https://acta.bibl.u-szeged.hu/75878/1/msznykonf_018_241-255.pdf
%http://acta.bibl.u-szeged.hu/75862/1/msznykonf_018_015-029.pdf

\section{Summarization dataset}\label{sec:dataset}
We built a dataset from Hungarian news articles using the Common Crawl corpus\footnote{https://commoncrawl.org/}. Our dataset can be used as training data for abstractive summarization and title generation tasks.

\subsection{Common Crawl}

The Common Crawl (CC) Foundation crawls web pages and provides periodic snapshots freely available to the public. The corpus contains petabytes of data collected since 2008. The data is available in WARC format on Amazon S3 and can be downloaded through HTTP or S3.  We used the CommonCrawl Downloader package \citep{indig_2018a}, which helped us download and deduplicate articles based on the given domains. We selected 9 large Hungarian news websites with plenty of content and downloaded pages from all Common Crawl segments till 2022-27.
Our selection was limited to websites that use dedicated lead fields.
The portals' topics cover politics, economics, sports and several others.
We collected around 94 GB of raw HTML data.

\subsection{Dataset}

We extracted the title, the lead, the article, the date of its creation and the tags, if it had any from each article.
We skipped articles if we did not find a lead during parsing.
We removed every link, embedded social media, recommendations and image captions from each article that would not help the model in the summarization task. Articles that are part of a live blog are dropped, as they are too short and often cannot be considered as an independent article without the rest of the blog. We also dropped galleries because they usually do not have meaningful content without the images.

After scraping the pages we had about 1.88 million articles. However, we filtered the data to discard articles that would not be valid input for a summarization model (e.g. too short article). Table~\ref{Tab:preprocess} shows the constraints that were applied to the length of the lead and the article. For word and sentence tokenization we used the \texttt{quntoken} package\footnote{https://github.com/nytud/quntoken}. We also removed articles where the lead was longer than the article's text, this removed roughly 17k articles. The constraint about minimum and maximum article characters removed 24k articles. After filtering for maximum number of lead sentences and minimum number of lead tokens, 10k and 25k articles were removed respectively. The largest number of articles were dropped when we filtered for the minimum sentence number of the article text, this filtering step removed roughly 300k articles. We also removed 3k non-Hungarian articles, which we identified using a language detection model, FastText \citep{joulin2016bag, joulin2016fasttext}.

\begin{table}[!htbp]
\centering
  \begin{tabular}{l r}
    \toprule
    \textbf{Constraint}  & \textbf{Value} \\
    \midrule
    min article characters                   & 200 \\
    max article characters                   & 15000 \\
    min lead tokens                     & 6 \\
    max lead sentences                  & 5 \\
    min article sentences               & 6 \\
    \bottomrule
  \end{tabular}
  \caption{Length constraints of the article and lead.}
  \label{Tab:preprocess}
\end{table}

After filtering, we deduplicated the articles using Locality Sensitive Hashing (LSH) to remove redundant articles. Two articles were considered equivalent if their article texts were 45\% similar. When two articles were identified similar, the one without lead was dropped, if both of them had lead we kept the one with later crawl time. With this threshold about 363k articles were removed. The deduplicated dataset consists of 1.14M data samples. Table~\ref{Tab:data_per_site} shows the number of samples from each news site.

\begin{table}[!htbp]
\centering
  \begin{tabular}{l r}
    \toprule
    \textbf{Site}  & \textbf{Count} \\
    \midrule
    24.hu                   &  359.4k \\
    origo.hu                     & 305.0k \\
    hvg.hu                     & 216.8k \\
    index.hu                  & 154.5k \\
    nepszava.hu              & 56.7k \\
    portfolio.hu              & 22.6k \\
    m4sport.hu              & 17.9k \\
    metropol.hu              & 11.1k \\
    telex.hu              & 4.4k \\
    \bottomrule
  \end{tabular}
  \caption{Number of data samples per site.}
  \label{Tab:data_per_site}
\end{table}

\begin{figure}[!htbp]
    \includegraphics[width=0.99\textwidth]{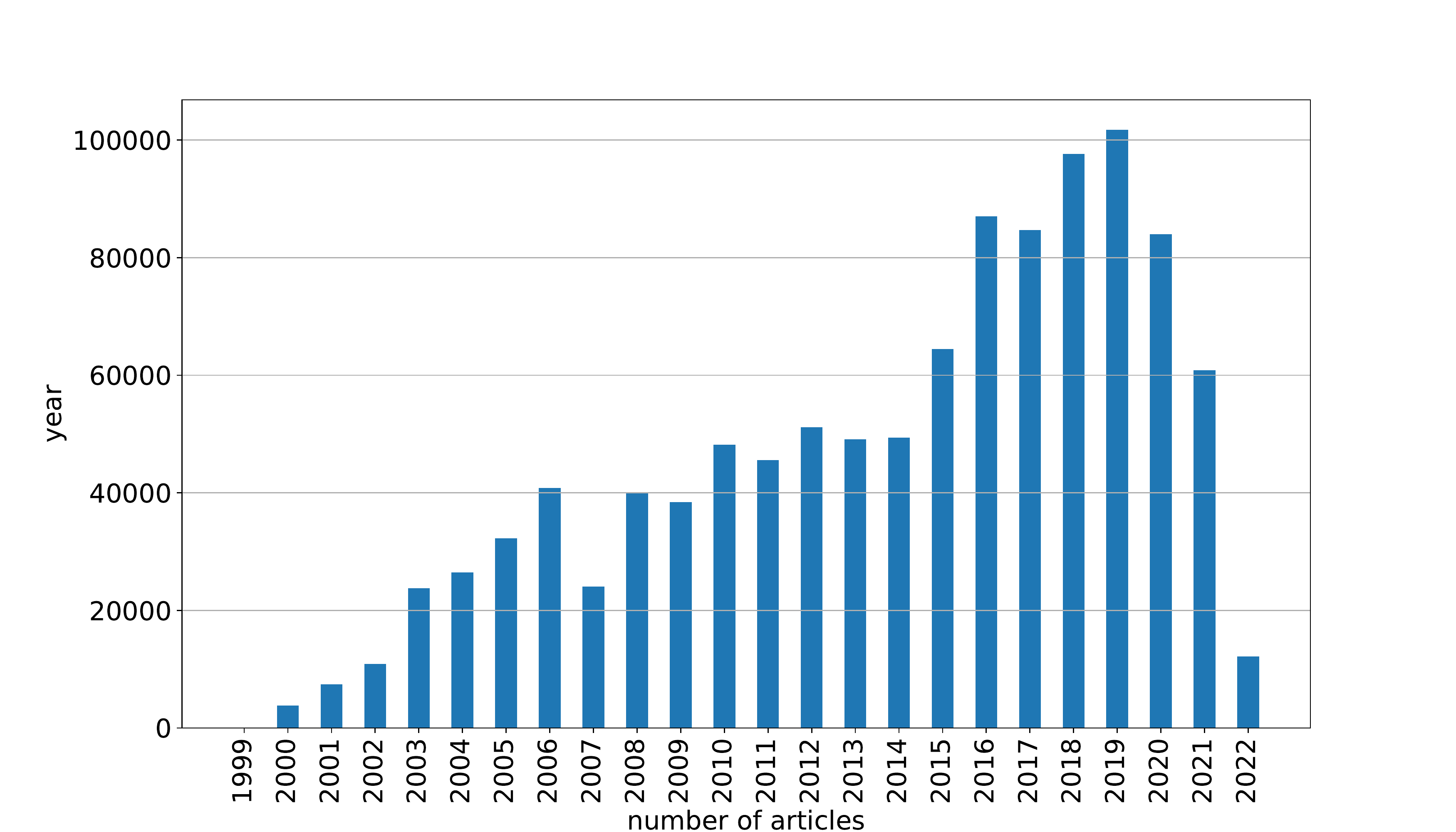}
    \caption{Number of articles by year.}
    \label{fig:article-by-year}
\end{figure}

The dataset contains articles from every year between 1999 and 2022. Figure~\ref{fig:article-by-year} shows the distribution of the articles among the years. The dataset has significantly fewer samples from 2022 despite using all Common Crawl segments up until July, 2022.
This is probably caused by the fact that CC has some latency.

Table~\ref{Tab:data_stats_per_site} describes the average number of tokens and sentences in the articles and leads. \textit{Népszava} and \textit{Telex} has significantly longer articles than the other sites.  

\begin{table}[!htbp]
\centering
  \begin{tabular}{l r r r r r r}
    \toprule
    \textbf{}  & \multicolumn{2}{c}{Article} & \multicolumn{2}{c}{Lead} \\
    \cmidrule(lr){2-3}                  
    \cmidrule(lr){4-5}
    \textbf{Site}  & \textbf{tokens} & \textbf{sents} & \textbf{tokens} & \textbf{sents} \\
    \midrule
    24.hu   & 323 & 17 & 21 & 1.3 \\
    origo.hu  &  396 & 19 & 40 & 1.9  \\
    hvg.hu    & 378 & 17 & 30 & 1.5 \\
    index.hu    & 517 & 26 & 43 & 2.2 \\
    nepszava.hu  & \textbf{869} & \textbf{39} & 32 & 1.5 \\
    portfolio.hu   & 457 & 22 & \textbf{52} & 2.1 \\
    m4sport.hu   & 385 & 23 & 28 & 1.3 \\
    metropol.hu  & 305 & 17 & 25 & 1.4 \\
    telex.hu    & 839 & 38 & 49 & \textbf{2.3} \\
    \bottomrule
  \end{tabular}
  \caption{Average length of the articles and leads.}
  \label{Tab:data_stats_per_site}
\end{table}

We compute a number of automatic quantitative metrics to estimate the quality of \textit{HunSum-1}. Novel n-gram ratio (NNG-n) \citep{narayan-etal-2018-dont} measures the percentage of n-grams in the summary that were not present in the original article. 
%\cite{bommasani-cardie-2020-intrinsic} generalize the abstractivity metric (ABS) originally proposed by \cite{grusky-etal-2018-newsroom}, where fragments are used to greedily match text spans between the article and the summary.
\cite{bommasani-cardie-2020-intrinsic} introduce a compression metric (CMP), which is defined as:
\begin{equation}
\begin{aligned}
\label{eq:cmp}
    \boldsymbol{CMP}(A, S)  = 1 - \frac{\left|S\right|}{\left|A\right|}
\end{aligned}
\end{equation}
where $\left|S\right|$ and $\left|A\right|$ are the length of the summary and the article. Finally, we compute a redundancy metric (RED-n) based on \cite{xlsum}. It is defined as a ratio of redundant n-grams and the total number of n-grams in a summary S:
\begin{equation}
\begin{aligned}
\label{eq:red}
    \boldsymbol{RED}(S)  = \frac{\sum_{i=1}^{m}(f_i - 1)}{\sum_{i=1}^{m}f_i} \\
    = 1 - \frac{m}{\left|S\right| - n + 1}
\end{aligned}
\end{equation}
where $\{f_1, f_2, ..., f_m\}$ denote the frequencies corresponding to unique n-grams $\{g_1, g_2, ..., g_m\}$. A summary is generally considered to be of high quality if it has a high novel n-gram ratio and compression with a low redundancy metric. The above metrics computed on the \textit{HunSum-1} dataset can be seen in Table~\ref{table:automatic-metrics}.

\begin{table}[!htbp]
  \centering
    \begin{tabular}{ c | c | c | c | c | c }
    \hline
    \textbf{NNG-1} & \textbf{NNG-2} & \textbf{NNG-3} & \textbf{CMP} & \textbf{RED-1} & \textbf{RED-2} \\
    \hline
    78.0 & 90.09 & 95.48 & 88.58 & 6.58 & 0.1 \\
    \hline
    \end{tabular}
  \caption{Intrinsic evaluation of \textit{HunSum-1}.}
  \label{table:automatic-metrics}
\end{table}

We created a train-dev-test split which will be available on the HuggingFace Dataset Hub under the name of \texttt{SZTAKI-HLT/HunSum-1}. The development and the test datasets both contain 2000 documents and are created using stratified sampling from every news page.

\section{Models}\label{sec:models}
We experimented with two language models: an encoder-decoder model, where  both sides are composed of BERT models \citep{bert2bert}.
We initialized both sides with the weights of the Hungarian version of BERT \citep{bert}, huBERT \citep{Nemeskey:2020} (Bert2Bert) and the pretrained multilingual mT5 \citep{mt5}. %In this section we give a brief introduction about these models.

\subsection{mT5}
mT5 is the multilingual version of the Text-To-Text Transfer Transformer (T5) \citep{t5} and it supports 101 languages including Hungarian. mT5 was trained on the mC4 dataset, totaling 6.3T tokens that was extracted from the Common Crawl dataset. Hungarian is the 22th most common language in this corpus and contains 37 million Hungarian pages. Unlike the T5 model, the mT5 was not trained on any downstream tasks, like abstractive summarization. mT5 has 5 variants: \texttt{mt5-small}, \texttt{mt5-base}, \texttt{mt5-large}, \texttt{mt5-xl} and \texttt{mt5-xxl}, with 300M, 580M, 1.2B, 3.7B and 13B parameters respectively. We decided to only experiment with the small and base variants due to computational constraints.
Unlike the BERT family, T5 and mT5 are generative models.

\section{Experiments}
We trained 3 summarization models: Bert2Bert, mT5-small and mT5-base. We found that different generation parameters work best for the Bert2Bert and the mT5 models. The mT5 models also needed an additional parameter, because we observed that the mT5 models tended to copy the input sentences. To prevent this behaviour, we set the \verb|encoder_no_repeat_ngram_size| argument to 4. Table~\ref{Tab:hyperparams} describes the training and the generation hyperparameters.

\begin{table}[!htbp]
\centering
  \begin{tabular}{l r r r}
    \toprule
    \textbf{Parameter}  & \textbf{Bert2Bert}  & \textbf{mT5-small} & \textbf{mT5-base} \\
    \midrule
    batch size & 13  & 16  &  12 \\
    learning rate & 5e-5  & 5e-5  &  5e-5 \\
    weight decay & 0.01  & 0.01  &  0.01 \\
    warmup steps & 16000  & 3000  &  3000 \\
    epochs & 15  & 10  &  10 \\
    \midrule
    \verb|no_repeat_ngram_size| & 3 & 3 & 3 \\
    \verb|num_beams| & 5 & 5 & 5 \\
    \verb|early_stopping| & True & False & False \\
    \verb|encoder_no_repeat_ngram_size| & - & 4 & 4 \\
    \bottomrule
  \end{tabular}
  \caption{Hyperparameters for training and generation.}
  \label{Tab:hyperparams}
\end{table}

%The following generating strategy was used for bert2bert: \{no\_repeat\_ngram\_size=3, num\_beams=5, early\_stopping=True\}. The mT5 models used the same parameters, except for the early\_stopping that was not set in this case. The mT5 models tended to copy the input sentences, so we set the encoder\_no\_repeat\_ngram\_size argument to 4 to prevent this behaviour. \todo{}

The maximum length of the article is 512 subwords and for the generated lead it is 128. The models were trained on a single NVIDIA A100 GPU. The Bert2Bert was trained for 265 hours, the mT5-small was trained for 54 hours, while the mT5-base model took 119 hours. Our models are available on the Huggingface Model Hub under the names of \texttt{SZTAKI-HLT/Bert2Bert-HunSum-1}, \texttt{SZTAKI-HLT/mT5-small-HunSum-1} and \texttt{SZTAKI-HLT/mT5-base-HunSum-1}.

\section{Results}

We used the ROUGE scores (Recall-Oriented Understudy for Gisting Evaluation) \citep{lin-2004-rouge} for automatic evaluation.
ROUGE takes a reference summary and computes the word ngram overlaps with the models' output.
The reference summary in our case is the real article lead.
We calculate two versions, a normal and a stemmed one.
As Hungarian is an agglutinative language, small changes in inflection can decrease the ROUGE scores even if the summary is correct.
We used the HuSpaCy \citep{HuSpaCy:2021} model for stemming. 
Table~\ref{Tab:results} shows the ROUGE scores on the test set.

\begin{table}[!htbp]
\centering
\setlength{\cmidrulekern}{0.25em}
  \begin{tabular}{l r r r r r r}
    \toprule
    \multirow{2}{*}{\textbf{Model}} & \multicolumn{3}{c}{Original} & \multicolumn{3}{c}{Stemmed} \\
    \cmidrule(lr){2-4}                  
    \cmidrule(lr){5-7}
    & \textbf{R-1}  & \textbf{R-2} & \textbf{R-L} & \textbf{R-1}  & \textbf{R-2} & \textbf{R-L} \\
    \midrule
    Bert2Bert & 28.52  & 10.35  & 20.07 & 31.75  & 11.80  &  22.31 \\
    mT5-small & 36.49 & 9.50  & 23.48 & 40.86  & 11.71  &  26.25 \\
    mT5-base  & \textbf{37.70} & \textbf{11.22} & \textbf{24.37} & \textbf{41.97}  & \textbf{13.65}  &  \textbf{27.23} \\
    \midrule
    foszt2oszt\tablefootnote{https://huggingface.co/BME-TMIT/foszt2oszt} & 23.62 & 5.05  & 16.57 & 26.22  & 5.95  &  18.37 \\
    hi-bart-base\tablefootnote{https://huggingface.co/NYTK/summarization-hi-bart-base-1024-hungarian} & 29.03 & 8.79 & 19.60 & 32.57 & 10.45 & 21.98 \\
    hi-bart\tablefootnote{https://huggingface.co/NYTK/summarization-hi-bart-hungarian} & 30.03 & 8.96 & 20.50 & 33.23  & 10.45  &  22.71 \\
    nol-bart\tablefootnote{https://huggingface.co/NYTK/summarization-nol-bart-hungarian} & 26.38 & 6.92 & 18.05 & 29.25 & 8.06 & 20.09 \\
    \bottomrule
  \end{tabular}
  \caption{ROUGE recall scores on the test set. ROUGE-1, ROUGE-2 and ROUGE-L scores are abbreviated as R-1, R-2 and R-L respectively.}
  \label{Tab:results}
\end{table}

Based on the ROUGE scores, the mT5-base model is clearly the best model and mT5-small is the second best, although not by every measure.
Despite its name, mT5-small is not a very small model with its 300M trainable parameters.
In contrast, Bert2Bert, which is the combination of two BERT models, only has 250M parameters.
Aside from the higher parameter count, T5 models are generative, while BERT is not, which may explain why a multilingual model is better for this generative task than the monolingual huBERT.

We compared our models to the available Hungarian summarization models. The results can be seen in Table~\ref{Tab:results}. Based on the scores on our test set, our mT5-base model outperformed the other Hungarian models.

%The Bert2Bert model achieved higher ROUGE-2 scores than the mT5-small despite having less parameters: Bert2Bert has roughly 250M parameters while the mT5-small has 300M. The mT5 models have significantly better results on the stemmed version. During the experiments, we observed that the mT5 models tend to output summaries that contains grammatical errors, especially related to inflection.

\subsection{Human evaluation}
As the ROUGE scores do not always represent the quality of the generated summaries, we decided to evaluate the models manually.
We sampled 80 articles from the test set with stratified sampling for human evaluation.
We used the evaluation framework from \cite{xlsum} (\textbf{Property A} to \textbf{C}), where an evaluator has to answer the following questions with \texttt{Yes} or \texttt{No}. We also added a new question about grammatical correctness (\textbf{Property D}).
\begin{itemize}
    \item \textbf{Property A:} Does the summary convey what the article is about?
    \item \textbf{Property B:} If the answer to property A is \texttt{Yes}, does the summary only contain information that is consistent with the article?
    \item \textbf{Property C:} If the answer to property A is \texttt{Yes}, does the summary only contain information that can be inferred from the article?
    \item \textbf{Property D:} Is the summary grammatically correct?
\end{itemize}

The 3 generated leads were presented to the annotators in random order for each article and the model name was hidden.
Every article was evaluated by three annotators, we computed the final answers by majority voting.

\begin{figure}[!htbp]
    \includegraphics[width=0.99\textwidth]{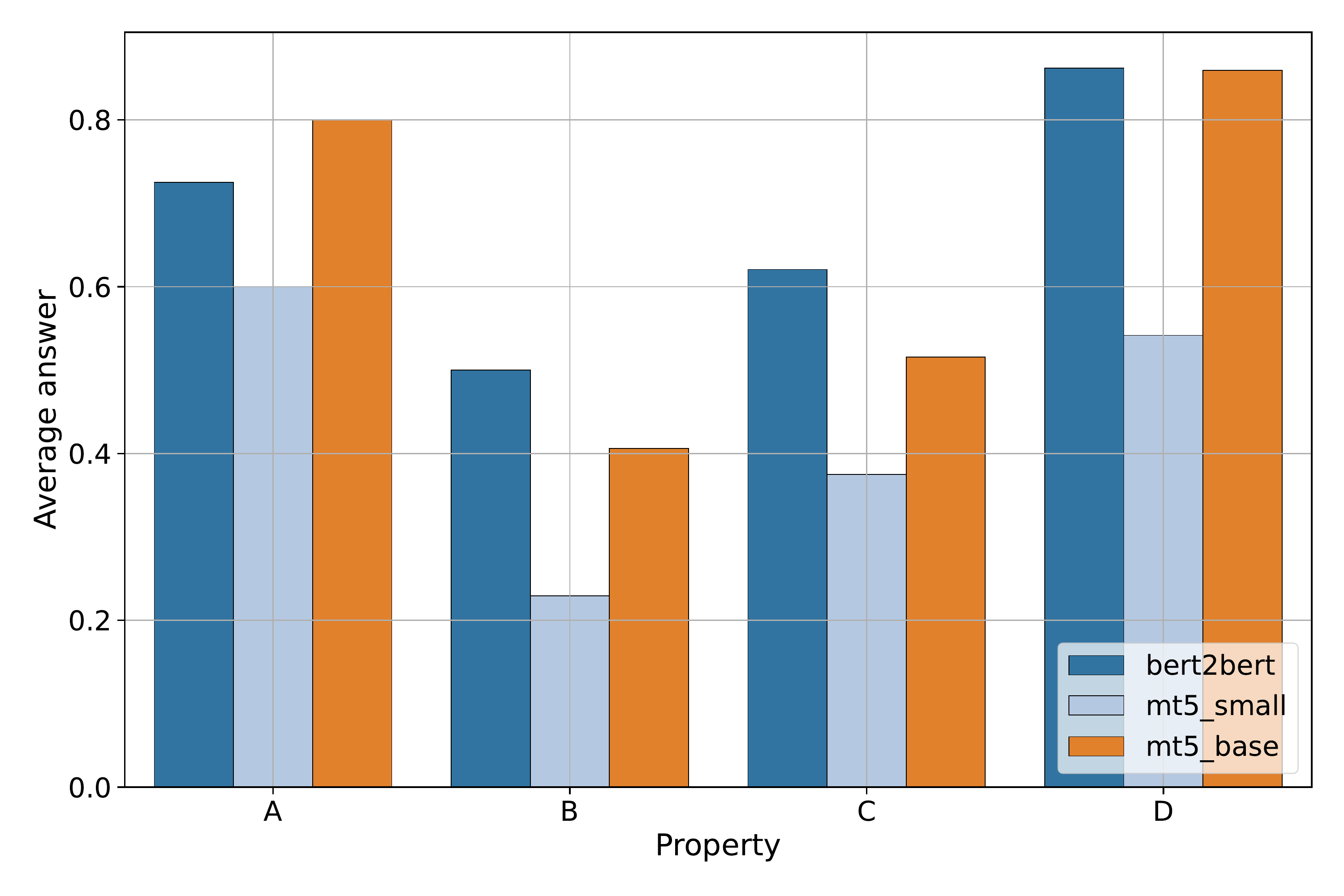}
    \caption{The average answers for the properties by models.}
    \label{fig:qualitative_eval_results}
\end{figure}

We took the average of the answers for every property-model pair, where 1 means \texttt{Yes} and 0 means \texttt{No} answer, Figure~\ref{fig:qualitative_eval_results} shows the results of the human evaluation. According to the human annotators, the mT5-small model performs the worst among the three models.
We also observe frequent inflection errors in this model.
The mT5-base model gives the best summary based on the content selection, but gives inconsistent information with the article more often than the Bert2Bert model.
The latter hallucinates less than the other two models.
The average answer for the grammatical question is significantly higher than the others, which means our models rarely output grammatically incorrect summaries.

We used the Cohen kappa \citep{cohen_kappa} to measure how much the annotators agreed on each question. We calculated this value for each annotator pair and took the average of it. The average pair-wised Cohen kappa is 56.77\%. The grammatical question has the lowest score (37.99\%) because the guideline was not clear and some of the annotators considered unfinished sentences grammatically incorrect, while others did not.

\section{Conclusion}

We introduced \textit{HunSum-1}, a Hungarian abstractive summarization corpus with over 1.1M data samples drawn from 9 major news outlets. We computed various intrinsic measures on the corpus quality. We then trained 3 summarization models: a Bert2Bert model and two mT5 models. We found that the larger mT5 model performed best both by automatic metrics and by human evaluation.

%We plan to extend our summarization dataset with more news pages, and newer Common Crawl segments.
%
%We will also try to improve our models for example by finetuning the hyperparameters. Apart from that our future work includes trying out other models, like mBERT and other versions of the mT5. As our generation is not perfect yet, we plan to do more experiments with generating using different parameters. 

% ---- Bibliography ----
%
%kulon bibtex fajl hasznalata
\renewcommand\bibname{\refname}
%magyar nyelvu cikkekhez
%\renewcommand\bibname{Hivatkoz\'asok}
%\bibliographystyle{splncsnat}
%for papers written in English
\bibliographystyle{splncsnat_en}
\bibliography{mszny}

\end{document}